# CATEGORICAL DATA AS A *STONE GUEST* IN A DATA SCIENCE PROJECT FOR PREDICTING DEFECTIVE WATER METERS

Giovanni Delnevo, Marco Roccetti, Luca Casini
Department of Computer Science and Engineering
Alma Mater Studiorum – University of Bologna
Via Zamboni, 33 – 40126 Bologna,
Italy
E-mail: {giovanni.delnevo2, marco.roccetti, luca.casini7}@unibo.it

**KEYWORDS**

Deep Learning, Human-Machine-Bigdata Interaction Loop, Human-in-the-Loop Methods, Defective Water Meters.

**ABSTRACT**

After a one-year long effort of research on the field, we developed a machine learning-based classifier, tailored to predict whether a mechanical water meter would fail with passage of time and intensive use as well. A recurrent deep neural network (RNN) was trained with data extrapolated from 15 million readings of water consumption, gathered from 1 million meters. The data we used for training were essentially of two types: *continuous* vs *categorical*. Categorical being a type of data that can take on one of a limited and fixed number of possible values, on the basis of some qualitative property; while continuous, in this case, are the values of the measurements. taken at the meters, of the quantity of consumed water (cubic meters). In this paper, we want to discuss the fact that while the prediction accuracy of our RNN has exceeded the 80% on average, based on the use of continuous data, those performances did not improve, significantly, with the introduction of categorical information during the training phase. From a specific viewpoint, this remains an unsolved and critical problem of our research. Yet, if we reason about this controversial case from a data science perspective, we realize that we have had a confirmation that accurate machine learning solutions cannot be built without the participation of domain experts, who can differentiate on the importance of (the relation between) different types of data, each with its own sense, validity, and implications. Past all the original hype, the science of data is thus evolving towards a multifaceted discipline, where the designations of data scientist/machine learning expert and domain expert are symbiotic.

**INTRODUCTION**

In the unfortunate event that you are in Copenhagen and need an ambulance, be aware that your conversation with the human operator of the call center will be listened by an Artificial Intelligence (AI) agent. It is probably *Corti*, an AI assistant that analyses the content and the tones of the conversation, together with the background noises, to detect heart attacks (Kreutzer and Sirrenberg 2019). This is only one of the many impressive tasks that AI systems have proven capable of, like beating the 99.8% of all human players at StarCraft 2 (The AlphaStar team 2019) or generating patrol routes to counteract poachers for wildlife protection, for example (Gholami et al. 2018).

However, the opacity of such algorithms, which are often referred to as *black boxes*, makes it difficult to understand what was actually learned and from which data (Burrell 2016), posing serious questions about their use (Shadowen 2019). To advance of a single step further in this direction, one could suppose that the introduction of qualitative contextual information (e.g., under the form of categorical data) could be put to good use to build more easily interpretable learning models.

Our experience here has been controversial. In this paper, in fact, we present the final phase of one research devoted to the design and development of a machine learning model able to predict the failure of a mechanical water meter. In previous phases, we adopted a specific methodology employed to extrapolate from the initial dataset of 15 million water meter readings only those examples that truly represent the complex phenomenon under observation. With that approach, we were able to train a Recurrent Neural Network (RNN) up to a prediction accuracy of more than 80% (Casini et al. 2019; Roccetti et al. 2019a; Roccetti et al. 2019b). Instead, what we discuss here is the surprising fact that adding categorical information to the numerical values of the readings did not contribute to the design of a better model. Results show that the addition of categorical variables, describing some qualitative properties of water meters, in fact, did not improve the prediction capabilities of the deep learning model we developed.

The remainder of this paper is structured as follows. In next Section, we illustrates the dataset, the preprocessing activities designed to extrapolate only safe and valid training examples, and the deep learning models we designed. The Section termed Results presents the accuracy performances of our models trained with (and without) categorical variables. The final Section concludes the paper.

**METHODS**

In this Section we describe the dataset, all the preprocessing activities that have preceded the training phase, and the deep learning models developed for this research.

**Dataset Description**

A company that distributes water in Northern Italy provided us with a dataset, consisting of over fifteen million water meter readings, relative to one million mechanical water meters.
On the categorical side, different attributes were available for each water meter. Among the others, the most relevant ones were: Producer, Meter Type, Year of Construction, and Type of Contract (signed by the client/consumer). Further, each reading was accompanied by the numerical value relative to the water consumed.

**Dataset Preprocessing**

A detailed description of both the preprocessing activities and the semantics of validity, defined with the domain experts of the company, employed to clean and prepare the numerical data, can be found in (Roccetti et al. 2019b). To summarize all thet work, it is enough to say here that the numerical value relative to a reading was considered valid only if:
- The reading had successfully met all the process requirements.
- The time, when the reading was taken, was considered *congruent*.
- The separation in time between two consecutive readings could not exceed a given predefined period (seven months).

Past several research phases at the end of which a subset of adequate water meters were selected for training, we built a number of positive/negative examples (i.e., defective/non defective water meters), as shown in Table 1 below, where added is also the correspondence between meter and valid readings available.

Table 1: Faulty/Non-faulty water meters (with at least *N* readings)

| # Readings | # Water Meters |
|---|---|
| 2 | 2279/18949 |
| 3 | 2062/15652 |
| 4 | 1912/12395 |
| 5 | 1736/10854 |

As to defective meters, an important role in the construction of our model is played by what we termed a *plateau*. Typically, when a meter is going to fail, for some reason, two or more consecutive values of consumed water are read that report exactly the same numerical value. This is a *plateau*. An intelligent system should predict a defective water meter before a plateau occurs.
For example, in Table 2 we show the number of defective water meters, whose associate readings are: i) the first reading of a plateau, plus the preceeding one (**1P+1**), ii) the first reading of a plateau, plus the two preceeding ones (**1P+2**), and so on …; and a) the first two readings of a plateau, plus the preceeding one (**2P+1**), ii) the first two reading of a plateau, plus the two preceeding ones (**2P+2**), and so on …

Table 2: Defective Water Meters with relative readings (including 1 or 2 plateau readings)

| #Readings | # Water Meters | #Readings | # Water Meters |
|---|---|---|---|
| 1P + 1 | 2279 | - | - |
| 1P + 2 | 2062 | 2P + 1 | 2260 |
| 1P + 3 | 1912 | 2P + 2 | 2055 |
| 1P + 4 | 1736 | 2P + 3 | 1905 |

**Deep Learning Models**

We developed two different learning models. The former (DNN1) takes as input only the numerical values associated to the readings; the latter (DNN2) takes as input both the numerical value associated to a reading, plus some categorical variables.
These two deep learning models have the following architectures. DNN1 is composed by three hidden layers: i) a Gated Recurrent Units, ii) a densely-connected layer with 32 neurons, and iii) a densely-connected layer with 128 neurons (Cho et al. 2014). Instead, DNN2 is comprised of two parallel sub-networks. The first subnetwork presents the same first two layers of DNN1 (i.e., those indicated with i) and ii)). The second subnetwork is composed by two densely-connected layers with respectively 128 and 96 neurons. The output of these two subnetworks is then merged and connected to a densely-connected layer with 128 neurons, that finally precedes the output layer. All the employed layers use a REctified Linear Unit (RELU) function to be activated (Glorot et al 2011).

**RESULTS**

The two models described above have been implemented with the Keras framework, utilizing Tensorflow as a backend. Both of them have been trained for eighty epochs, using *Adam* as the optimization algorithm (Kingma and Ba 2015). As the evaluation metrics, we adopted the Area Under the Curve (AUC) of the Receiver Operating Characteristic (ROC) (Hanley and McNeil 1982). For each experiment, we have used 80% of the available examples (i.e., water meters) for the training phase, while the remaining 20% was used for the testing phase. In the training phase, we anticipated some of the testing with a traditional ten-fold cross-validation technique.

**Without Categorical attributes**

We have trained and tested our DNN1 model using only the numerical values associated to the reading. The results with the **1P** situation are reported in Table 3. As shown, the ability of our model to predict whether a meter is defective is good both in the validation and testing phases. The adequacy of this model is confirmed by the good results that can be achieved even in the **2P** case (around 90%), not shown here for the sake of conciseness.

Table 3: Cross-validation and testing results (AUC) using
**1P-type** readings, **without** categorical attributes

| Readings | Cross-validation | Testing |
|---|---|---|
| 1P + 1 | 85% | 84% |
| 1P + 2 | 85% | 86% |
| 1P + 3 | 84% | 84% |
| 1P + 4 | 84% | 86% |

**With Categorical attributes**

The DNN2 model was instead trained using both the continuous values and some categorical attributes among those mentioned at the beginning of this paper. Surprising are the results shown in Table 4 (under the **1P** situation). In fact, in all examined cases, the use of categorical data does not improve the prediction capabilities of the DNN2 model with respect the DNN1 one.

Table 4: Cross-validation and testing results (AUC) using
**1P-type** readings, **with** categorical attributes

| Readings | Cross-validation | Testing |
|---|---|---|
| 1P + 1 | 77% | 81% |
| 1P + 2 | 77% | 83% |
| 1P + 3 | 72% | 83% |
| 1P + 4 | 74% | 83% |

For now, the lack of transparency of the deep neural models have made it impossible to know the reasons behind this behavior. At the current stage of our research, we could only provide hypothesis to explain this surprising result, yet we do not have enough data to discuss this issue, at a deeper level of comprehension. We would like to quit this paper, hence, simply emphasizing that these non positive results could be a further confirmation of the intuition that the role played by domain experts is crucial. Being them the only ones with the ability to differentiate on the importance of different types of data, each with its own sense, validity, and implications (Ansari et al. 2018; Holzinger et al. 2019).

**CONCLUSION**

We presented a further step in our experience in the design of a ML classifier that predicts water meters failures based on measurements of water consumptions. We evaluated the impact of categorical attributes, added to the readings to build a new DL model. Results show that the introduction of those categorical data did not improve the accuracy of the classification. This may be a confirmation of the importance of engaging domain experts in the design of a DL model, to differentiate on the importance of different types of data and of their adequate use to train an intelligent machine (Delnevo et al. 2019; Palazzi et al. 2004; Roccetti et al. 2010).